\documentclass[10pt,journal]{IEEEtran}
\usepackage{amsmath,amsfonts}
\usepackage{algorithmic}
\usepackage{array}
\usepackage[caption=false,font=footnotesize,labelfont=rm,textfont=rm]{subfig}
\usepackage{textcomp}
\usepackage{stfloats}
\usepackage{url}
\usepackage{verbatim}
\usepackage{graphicx}
\usepackage{xcolor}
\usepackage{caption}
\def\BibTeX{{\rm B\kern-.05em{\sc i\kern-.025em b}\kern-.08em
    T\kern-.1667em\lower.7ex\hbox{E}\kern-.125emX}}
\usepackage{balance}
\begin{document}
\title{Gaussian Combined Distance: A Generic Metric for Object Detection}

\author{
	\IEEEauthorblockN{
		Ziqian Guan\IEEEauthorrefmark{1}, 
		Xieyi Fu\IEEEauthorrefmark{1}, 
		Pengjun Huang\IEEEauthorrefmark{1}, 
		Hengyuan Zhang\IEEEauthorrefmark{1},
    Hubin Du\IEEEauthorrefmark{1},\\
    Yongtao Liu\IEEEauthorrefmark{1},
    Yinglin Wang\IEEEauthorrefmark{2},
    Qang Ma\IEEEauthorrefmark{2}}\\
	\IEEEauthorblockA{\IEEEauthorrefmark{1}North China Institute of Science and Technology\\}
	\IEEEauthorblockA{\IEEEauthorrefmark{2}Hegang Industrial Technology Service Co., Ltd}
  \thanks{Ziqian Guan, Xieyi Fu, Pengjun Huang, Hengyuan Zhang, Hubin Du, and Yongtao Liu are with the North China Institute of Science and Technology; Yinglin Wang and Qang Ma are with Hegang Industrial Technology Service Co., Ltd.}
} 

\maketitle

\begin{abstract}
In object detection, a well-defined similarity metric can significantly enhance model performance. Currently, the IoU-based similarity metric is the most commonly preferred choice for detectors. However, detectors using IoU as a similarity metric often perform poorly when detecting small objects because of their sensitivity to minor positional deviations. To address this issue, recent studies have proposed the Wasserstein Distance as an alternative to IoU for measuring the similarity of Gaussian-distributed bounding boxes. However, we have observed that the Wasserstein Distance lacks scale invariance, which negatively impacts the model's generalization capability. Additionally, when used as a loss function, its independent optimization of the center attributes leads to slow model convergence and unsatisfactory detection precision. To address these challenges, we introduce the Gaussian Combined Distance (GCD). Through analytical examination of GCD and its gradient, we demonstrate that GCD not only possesses scale invariance but also facilitates joint optimization, which enhances model localization performance. Extensive experiments on the AI-TOD-v2 dataset for tiny object detection show that GCD, as a bounding box regression loss function and label assignment metric, achieves state-of-the-art performance across various detectors. We further validated the generalizability of GCD on the MS-COCO-2017 and Visdrone-2019 datasets, where it outperforms the Wasserstein Distance across diverse scales of datasets.
Code is available at https://github.com/MArKkwanGuan/mmdet-GCD.
\end{abstract}

\begin{IEEEkeywords}
Tiny Object Detection, Generic Metric.
\end{IEEEkeywords}

\section{Introduction}
\label{introduction}

In recent years, deep neural networks have driven significant advancements in computer vision, particularly in object detection. However, most research has focused on standard-sized objects, 
overlooking the specific challenges associated with detecting small objects.
These challenges are crucial in many practical applications, including aerial photography\cite{xu2021dot}, large-scale surveillance, and maritime rescue\cite{yu2020scale}. 
Small objects in the AI-TOD\cite{wang2021tiny} dataset, typically smaller than $16 \times 16$ pixels, pose significant challenges for feature extraction and increase the likelihood of detection errors. Research on small 
object detection has primarily focused on multi-scale feature learning\cite{he2015spatial}\cite{lin2017feature}, data augmentation techniques\cite{bai2018sod}\cite{li2017perceptual}, and the design of backbone 
networks\cite{bell2016inside}\cite{ding2019learning} incorporating attention mechanisms. Although these strategies have improved detection performance, they often require higher computational costs to enhance 
accuracy and may compromise the detection of standard-sized objects.

Assessing bounding box similarity is vital for object detection models. Traditional similarity metrics, such as L$_n$-norms or Intersection over Union (IoU)\cite{yu2016unitbox}, are used. The latter has become more favored in 
recent detection methods due to its ability to consider the interrelation between bounding boxes, despite the critical flaw that IoU fails to provide sufficient gradients when there is no overlap. This issue has 
been partially addressed by variants such as GIoU\cite{rezatofighi2019generalized}, DIoU\cite{zheng2020distance}, and EIoU\cite{zhang2022focal}. However, small positional differences in small objects can significantly reduce IoU, impeding model optimization and resulting in slow convergence 
and inaccurate positioning. To overcome these issues, the Wasserstein Distance has been introduced as an alternative measure of bounding box similarity. Its primary advantage is its ability to assess distribution 
similarity, maintaining effectiveness even without overlap.

Despite the advantages of the Wasserstein Distance\cite{yang2021rethinking}, it is not scale invariant, which is a drawback for datasets with a wide range of object sizes. To address this issue, the Normalized 
Wasserstein Distance (NWD)\cite{xu2022detecting}\cite{xue2023visual} was proposed, incorporating a hyperparameter C that represents the average size of objects in the dataset. However, NWD's performance remains 
inconsistent across various general datasets. Therefore, we propose the Gaussian Combined Distance (GCD), a universal similarity metric. Due to its scale invariance, GCD improves the accuracy of detecting small 
targets while maintaining comparable performance on standard-sized datasets. The analysis of GCD and its gradients reveals its joint optimization properties during the learning process, leading to superior detection 
performance, particularly for small targets. The main highlights of this paper include:

\begin{itemize}
    \item We highlighted the absence of robust scale invariance in NWD and showed that its inherent feature of independently optimizing centrality results in reduced detection accuracy. Consequently, we proposed GCD as a metric for gauging the similarity between two bounding boxes.
    \item GCD possesses scale invariance and includes a jointly optimizing characteristic, significantly improving the precision of small object detection.
    \item As both a loss function and a label assignment metric, GCD significantly enhances the detection of small objects in bounding box regression-based detectors. It achieved state-of-the-art (SOTA) performance on the AI-TOD-v2\cite{xu2022detecting} dataset, while maintaining robust generalization across diverse datasets.
\end{itemize}

\section{Related Work}
\label{related work}
\subsection{Small Object Detection}
Recent studies have primarily focused on leveraging contextual information and attention mechanisms to enhance the detection capabilities of models for small objects\cite{lin2017feature}\cite{zhao2019m2det}. 
Multiscale learning, validated by prior research, effectively integrates features across various scales\cite{lin2017feature}, thereby improving target detection performance. 
Feature enhancement-based detectors aim to augment the feature representations of small objects\cite{singh2018analysis}\cite{singh2018sniper}, often employing super-resolution techniques or generative adversarial 
networks (GANs)\cite{bai2018sod}\cite{li2017perceptual} for this enhancement. Compared to our method, these approaches incur additional computational costs.

\subsection{Similarity Metric in Object Detection}
To mitigate the sensitivity of the L$_2$-norm to the size of bounding boxes, YOLOv1\cite{redmon2016you} introduced a square root transformation of bounding box dimensions to lessen the impact of larger boxes, while 
YOLOv3\cite{redmon2018yolov3} incorporated a penalty term to reduce their dominance. Fast R-CNN\cite{girshick2015fast} and Faster R-CNN\cite{ren2015faster} adopted the Smooth-L$_1$\cite{girshick2015fast} loss 
function, which is less sensitive to outliers compared to the L$_2$-norm. Unlike Intersection over Union (IoU), which overlooks the geometric correlation between bounding boxes, leading to suboptimal performance, 
GIoU\cite{rezatofighi2019generalized} introduced a penalty term constructed from the smallest enclosing box to address the issue of gradient vanishing when two bounding boxes do not overlap. DIoU\cite{zheng2020distance} employs a distance-based penalty term, and CIoU\cite{zheng2020distance} 
builds on DIoU\cite{zheng2020distance} by incorporating an aspect ratio measure. 
WD\cite{xu2022detecting} and KLD\cite{yang2021learning} were proposed based on Gaussian Bounding Boxes (GBBs), achieving superior results on TOD compared to IoU-based methods. However, they failed to demonstrate consistent performance across datasets of 
varying scales and even exhibited a certain degree of degradation on standard-scale datasets.

\section{Methodology}

\subsection{Gaussian Distribution Modeling}

The conventional representation of bounding boxes utilizes an axis-aligned rectangle, denoted by ($x$, $y$, $w$, $h$), where ($x$, $y$) represent the center coordinates, and $w$ and $h$ denote the width and height 
of the bounding box, respectively. For small objects, bounding boxes often encompass background pixels because real-world objects are seldom perfect rectangles. 
Within these bounding boxes, foreground pixels are concentrated near the center, while background pixels tend to be located towards the edges. 
To more accurately represent the varying significance of pixels within the bounding box, it is advantageous to model the bounding box as a two-dimensional (2D) Gaussian distribution, in which central pixels carry the highest weight, and the importance of pixels decreases radially from the center to the periphery.
Specifically, a two-dimensional Gaussian distribution, denoted as $\mathcal{N}(\mu, \Sigma)$, where $\mu$ and $\Sigma$ represent the mean vector and covariance matrix, respectively, can be expressed as:

\begin{equation}
    \label{eq:Gaussian}
    \mathbf{\mu}=\begin{bmatrix}
        x\\ y
    \end{bmatrix}
    ,
    \mathbf{\Sigma}=\begin{bmatrix}
    \frac{w^2}{4} & 0 \\ 
    0 & \frac{h^2}{4}
    \end{bmatrix}
\end{equation}

Therefore, the bounding box is represented by a Gaussian distribution.

\subsection{Gaussian Combined Distance}

Universal metrics employed in object detection generally must satisfy the following criteria:
\begin{itemize}
    \item \textbf{Criterion 1:} Affine invariance and symmetry.
    \item \textbf{Criterion 2:} Differentiability, avoiding vanishing or exploding gradients.
    \item \textbf{Criterion 3:} Smooth boundary processing.
\end{itemize}
Therefore, we propose constructing the GCD to simultaneously satisfy all these specified criteria.

The GCD between $\mathbf{X}_{p}$ and $\mathbf{X}_{t}$ is:
\begin{equation}
    \begin{aligned}
        \mathbf{D}_{gc}^2\left(\mathcal{N}_p, \mathcal{N}_t\right)&=({\mu}_{p}-{\mu}_{t})^{\top}2{\Sigma}_{p}^{-1}({\mu}_{p}-{\mu}_{t})\\
        &+({\mu}_{t}-{\mu}_{p})^{\top}2{\Sigma}_{t}^{-1}({\mu}_{t}-{\mu}_{p})\\
        &+2({\Sigma}_{p}^{-1/2})^{\top}\lVert\mathbf{\Sigma}_{p}^{1/2}-\mathbf{\Sigma}_{t}^{1/2}\rVert_{F}^{2}({\Sigma}_{p}^{-1/2})\\
        &+2({\Sigma}_{t}^{-1/2})^{\top}\lVert\mathbf{\Sigma}_{t}^{1/2}-\mathbf{\Sigma}_{p}^{1/2}\rVert_{F}^{2}({\Sigma}_{t}^{-1/2})\\
    \end{aligned}
    \label{eq:gcd}
\end{equation}
where $\left \| \cdot  \right \|_{F}$ is the Frobenius norm.

The Eq.~\ref{eq:gcd} can be simplified as:
\begin{equation}
    \label{eq:gcd_sim}
    \begin{aligned}
        \mathbf{D}_{gc}^2\left(\mathcal{N}_p, \mathcal{N}_t\right)&=\frac{1}{2}\left(\frac{(x_{p}-x_{t})^2}{w_{p}^2}+\frac{(y_{p}-y_{t})^2}{h_{p}^2}\right)\\
        &+\frac{1}{2}\left(\frac{(w_{p}-w_{t})^2}{4w_{p}^2}+\frac{(h_{p}-h_{t})^2}{4h_{p}^2}\right)\\
        &+\frac{1}{2}\left(\frac{(x_{t}-x_{p})^2}{w_{t}^2}+\frac{(y_{t}-y_{p})^2}{h_{t}^2}\right)\\
        &+\frac{1}{2}\left(\frac{(w_{t}-w_{p})^2}{4w_{t}^2}+\frac{(h_{t}-h_{p})^2}{4h_{t}^2}\right)
    \end{aligned}
\end{equation}

The GCD satisfies the symmetry property. We provide the following proof to demonstrate affine invariance:

For a full-rank matrix $\mathbf{M}$, $\lvert\mathbf{M}\rvert$ $\neq$ 0, we have ${\mathbf{D}_{gc}^2}(\mathcal{N}_{p}||\mathcal{N}_{t})$ $=$ ${\mathbf{D}_{gc}^2}(\mathcal{N}_{p^{'}}||\mathcal{N}_{t^{'}})$, 
$\mathbf{X}_{p^{'}}$ $=$ $\mathbf{M}\mathbf{X}_{p}$ $\sim$ $\mathcal{N}_{p}(\mathbf{M}{\mu}_{p},\mathbf{M}{\Sigma}_{p}\mathbf{M}^{\top})$, $\mathbf{X}_{t^{'}}$ $=$ $\mathbf{M}\mathbf
{X}_{t}$ $\sim$ $\mathcal{N}_{t}(\mathbf{M}{\mu}_{t},\mathbf{M}{\Sigma}_{t}\mathbf{M}^{\top})$, denoted as $\mathcal{N}_{p^{'}}$ and $\mathcal{N}_{t^{'}}$.

For the center distance term of $\mathbf{D}_{gc}^2(\mathcal{N}_{t^{'}}||\mathcal{N}_{p^{'}})$, the expression as follows:
\begin{equation}
    \begin{aligned}
        ({\mu}_{p}-{\mu}_{t})&^{\top}\mathbf{M}^{\top}(\mathbf{M}^{\top})^{-1}2{\Sigma}_{p}^{-1}\mathbf{M}^{-1}\mathbf{M}({\mu}_{p}-{\mu}_{t}) \\
        &=({\mu}_{p}-{\mu}_{t})^{\top}2{\Sigma}_{p}^{-1}({\mu}_{p}-{\mu}_{t})
    \end{aligned}
    \label{eq:proof_scale_invariance1}
\end{equation}

For the coupling term of $\mathbf{D}_{gc}^2(\mathcal{N}_{t^{'}}||\mathcal{N}_{p^{'}})$, the expression as follows:
\begin{equation}
    \begin{aligned}
        &2(\mathbf{M}^{-1/2})^{\top}({\Sigma}_{p}^{-1/2})^{\top}(\mathbf{M}^{-1/2})(\mathbf{M}^{1/2})\lVert\mathbf{\Sigma}_{p}^{1/2}\\
        &-\mathbf{\Sigma}_{t}^{1/2}\rVert_{F}^{2}(\mathbf{M}^{1/2})^{\top}(\mathbf{M}^{-1/2})^{\top}({\Sigma}_{p}^{-1/2})(\mathbf{M}^{-1/2})\\
        &=2({\Sigma}_{p}^{-1/2})^{\top}\lVert\mathbf{\Sigma}_{p}^{1/2}-\mathbf{\Sigma}_{t}^{1/2}\rVert_{F}^{2}({\Sigma}_{p}^{-1/2})\\
    \end{aligned}
    \label{eq:proof_scale_invariance2}
\end{equation}

It is evident that the subsequent terms of the GCD possess a similar structure. By employing a comparable simplification process, we obtain:
\begin{equation}
    \begin{aligned}
        \mathbf{D}_{gc}^2(\mathcal{N}_{p^{'}}||\mathcal{N}_{t^{'}})=\mathbf{D}_{gc}^2(\mathcal{N}_{p}||\mathcal{N}_{t})
    \end{aligned}
    \label{eq:proof_scale_invariance3}
\end{equation}

As shown in Eq.~\ref{eq:proof_scale_invariance3}, common metrics such as GCD and IoU demonstrate scale invariance, unlike WD\cite{xu2022detecting}. This lack of scale invariance 
significantly contributes to WD's diminished performance on extensive datasets.

IoU and KLD\cite{yang2021learning} exhibit zero gradients when the bounding box overlap is minimal, leading to insufficient supervision signals for small targets during model training. 
Although WD\cite{xu2022detecting} maintains non-zero gradients even when the bounding box overlap is small, it treats bounding boxes with different degrees of shift uniformly, 
thereby hampering the model's high-precision detection performance. In contrast, KLD\cite{yang2021learning} and GCD exhibit similar gradient curves. By assigning larger gradients 
to more accurately positioned bounding boxes, they enhance the model's high-precision detection performance.

\begin{equation}
    \begin{aligned}
        \frac{\partial \mathbf{D}_{gc}^2(\mu_{p})}{\partial\mu_{p}} =
        \begin{bmatrix}
        \frac{(w_{t}^{2}+w_{p}^{2})(x_{p}-x_{t})}{w_{t}^{2}w_{p}^{2}}\\
        \frac{(h_{t}^{2}+h_{p}^{2})(y_{p}-y_{t})}{h_{t}^{2}h_{p}^{2}}
        \end{bmatrix}
    \end{aligned}
    \label{eq:center_gradient}
\end{equation}

As shown in Eq.~\ref{eq:center_gradient}, the GCD optimizes the gradient of the center distance. 
The weights $\frac{1}{w_{t}^{2}w_{p}^{2}}$ and $\frac{h_{t}^{2} + h_{p}^{2}}{h_{t}^{2}h_{p}^{2}}$ enable 
the model to adjust the gradient dynamically according to scale during training. When the target size is 
small or the edges are short, even slight deviations in the corresponding direction can lead to significant 
changes in IoU and the related gradients, which is detrimental to training. Specifically, for these targets, 
the GCD increases the emphasis on optimizing in the corresponding direction by assigning larger gradient gains. 
In contrast, when optimizing the center distance using WD\cite{xu2022detecting}, the gradient is $\left(2(x_{p}-x_{t}), 2(y_{p}-y_{t})\right)^\top$. 
This indicates that WD's optimization of center distance is independent and does not adjust the gradient dynamically 
based on the width and height of the predicted box, weakening the model's high-precision detection performance.

\begin{equation}
    \begin{aligned}
        &\frac{\partial \mathbf{D}_{gc}^2({\Sigma}_{p})}{\partial {h_{p}}} = (h_{p}h_{t}-h_{t}^{2})(\frac{h_{t}^{3}+h_{p}^{3}}{4h_{t}^{3}h_{p}^{3}})-\frac{(y_{p}-y_{t})^2}{h_{p}^{3}} \\
        &\frac{\partial \mathbf{D}_{gc}^2({\Sigma}_{p})}{\partial {w_{p}}} = (w_{p}w_{t}-w_{t}^{2})(\frac{w_{t}^{3}+w_{p}^{3}}{4w_{t}^{3}w_{p}^{3}})-\frac{(x_{p}-x_{t})^2}{w_{p}^{3}}
        \label{eq:hw_gradient}
    \end{aligned}
\end{equation}

Eq.~\ref{eq:hw_gradient} illustrates how GCD optimizes gradients for both width and height. Similar to optimizing center distance, 
GCD enhances gradients for targets with smaller widths and heights, thereby emphasizing these targets more prominently. Concurrently, 
penalty terms associated with center distance control the asynchronous optimization of the center relative to width and height during training. 
When $w_p = w_t$ and $h_p = h_t$, the gradient for optimizing width and height simplifies to 
$\frac{(x_{p}-x_{t})^2}{-w_{p}^{3}}$ and $\frac{(y_{p}-y_{t})^2}{-h_{p}^{3}}$, respectively, while the gradient for optimizing center distance simplifies 
to $\left(\frac{2}{w_{t}^2}(x_{p}-x_{t}), \frac{2}{h_{t}^2}(y_{p}-y_{t})\right)^\top$. Therefore, GCD exhibits joint optimization characteristics similar to KLD\cite{yang2021learning} but avoids 
the issue of vanishing gradients inherent in logarithmic functions.

\subsection{Metric Normalization}

Given that the range of the GCD extends beyond $[0,1]$, directly utilizing it as a similarity measure may render it excessively 
sensitive to substantial errors. To mitigate this issue, we employ a nonlinear transformation to convert the GCD into a more refined and expressive 
metric, as demonstrated in Eq.~\ref{eq:norm}.

\begin{equation}
    \begin{aligned}
        \mathbf{M}_{gcd} = \exp \left(-{\sqrt{\mathbf{D}_{gc}^{2}\left(\mathcal{N}_p, \mathcal{N}_t\right)}}\right)
    \end{aligned}
    \label{eq:norm}
\end{equation}

Thus, the GCD satisfies all the criteria for a general metric.

\section{Experiments}

Our experiments encompass a diverse range of datasets, including AI-TOD-v2 \cite{xu2022detecting}, VisDrone-2019 \cite{du2019visdrone}, and 
MS-COCO-2017 \cite{lin2014microsoft}. To ensure a fair comparison of loss function performance, all ablation studies are performed using the 
MMDetection \cite{chen2019mmdetection} codebase. We consistently employ a ResNet-50 \cite{he2016deep} backbone network, pretrained on ImageNet 
\cite{russakovsky2015imagenet} and enhanced with a Feature Pyramid Network (FPN) \cite{lin2017feature}. Training spans 12 epochs, utilizing the 
SGD optimizer with a momentum of 0.9, a weight decay of $10^{-4}$, and a batch size of 8 for all datasets. For Faster R-CNN, we replaced the label 
assignment metrics and loss functions only at the RPN stage.The initial learning rate is set to 0.01 and is reduced by an order of magnitude at 
epochs 8 and 11. 

\begin{figure*}[t]
  \centering
  \includegraphics[width=0.18\paperwidth]{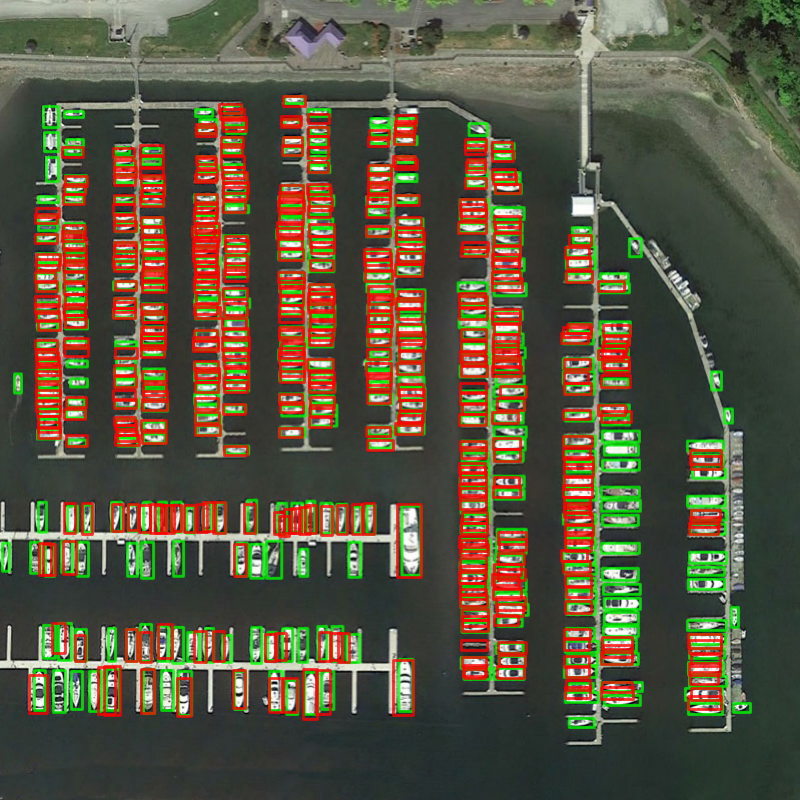}
  \includegraphics[width=0.18\paperwidth]{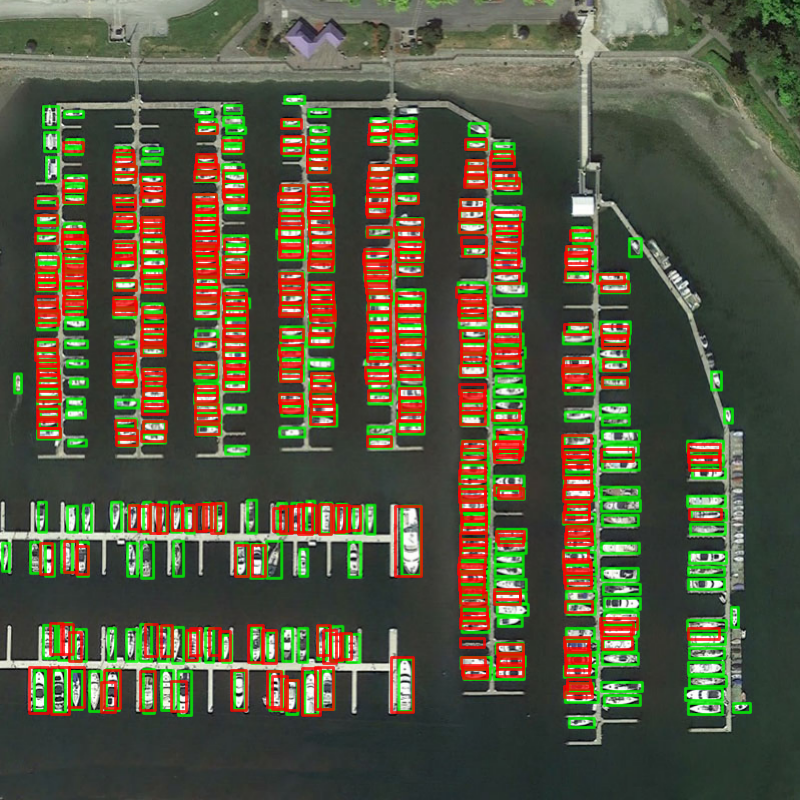}
  \includegraphics[width=0.18\paperwidth]{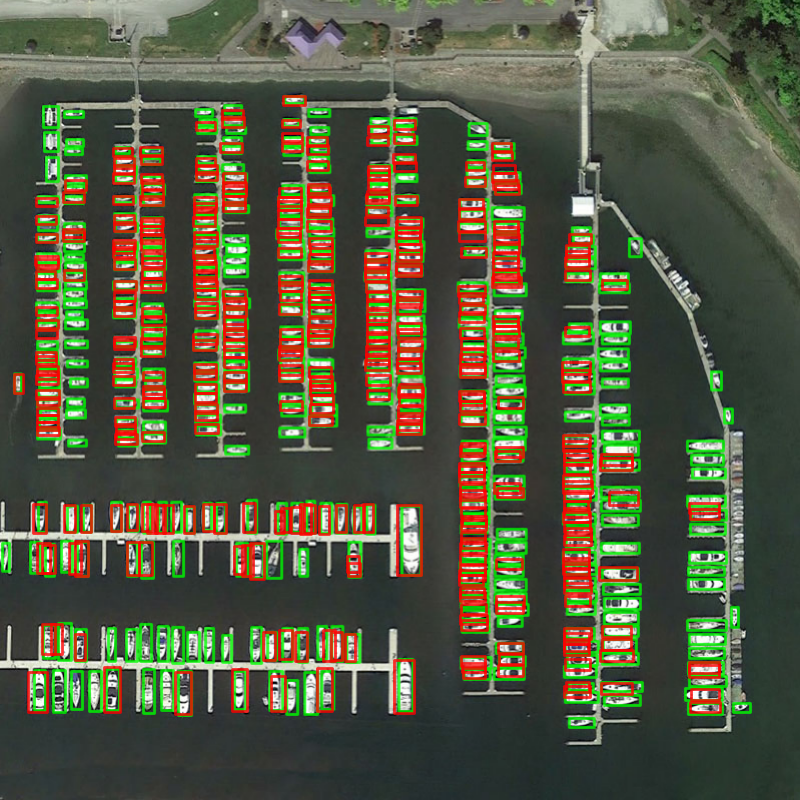}
  \includegraphics[width=0.18\paperwidth]{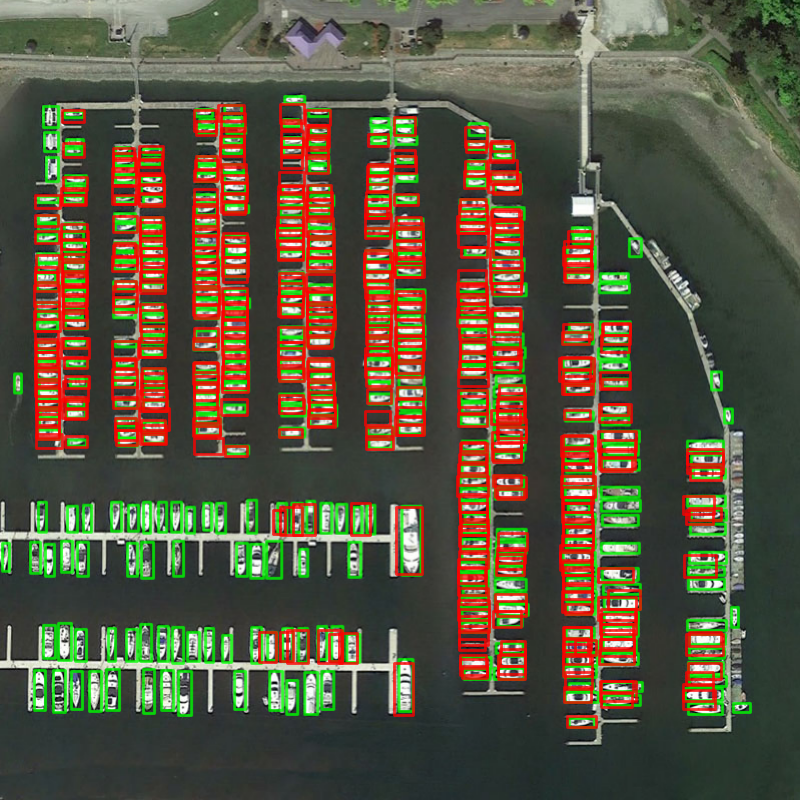}
  \includegraphics[width=0.18\paperwidth]{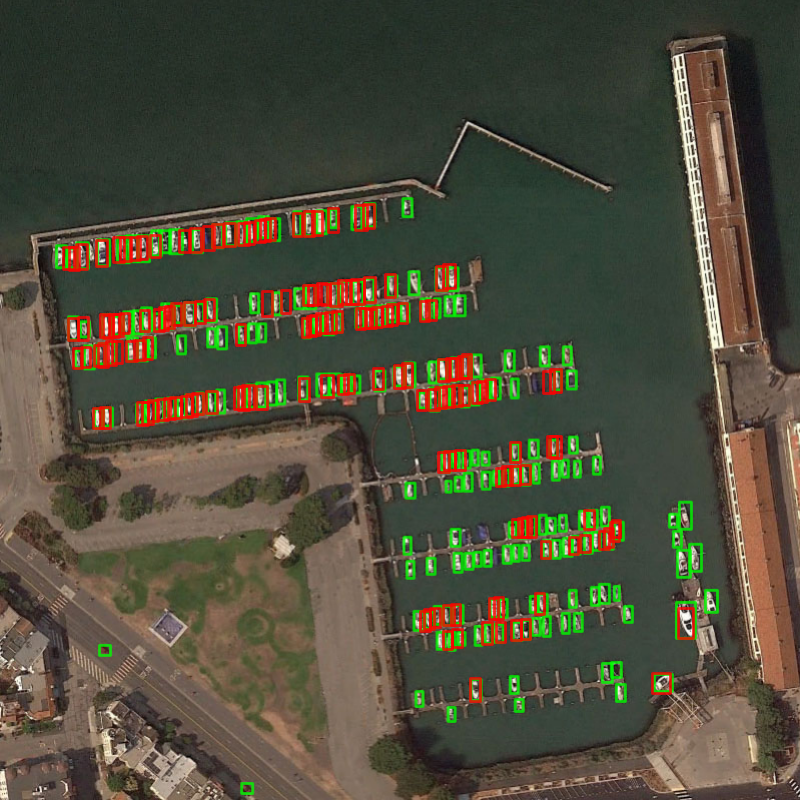}
  \includegraphics[width=0.18\paperwidth]{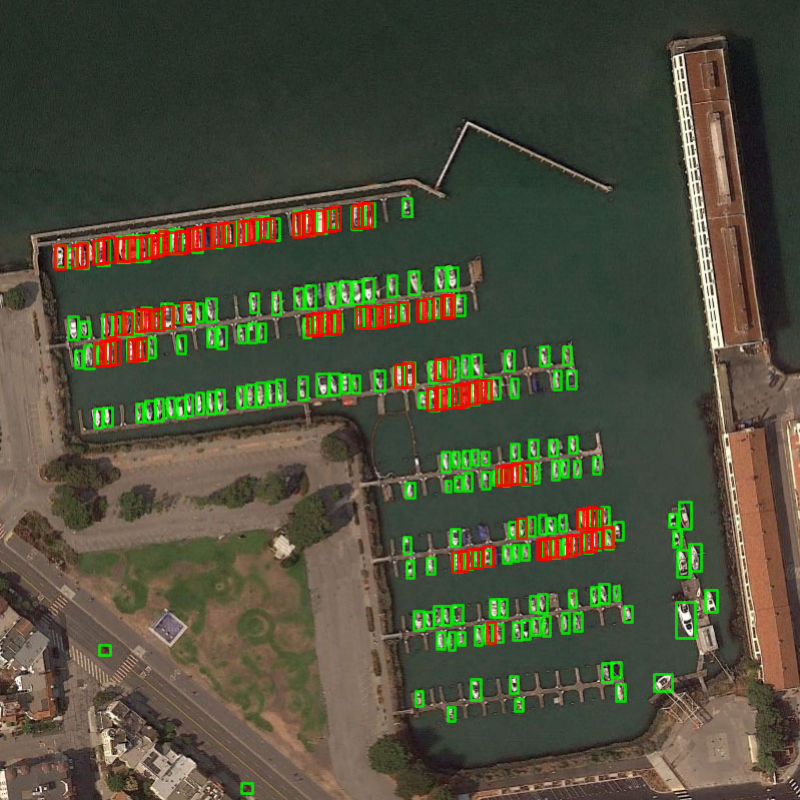}
  \includegraphics[width=0.18\paperwidth]{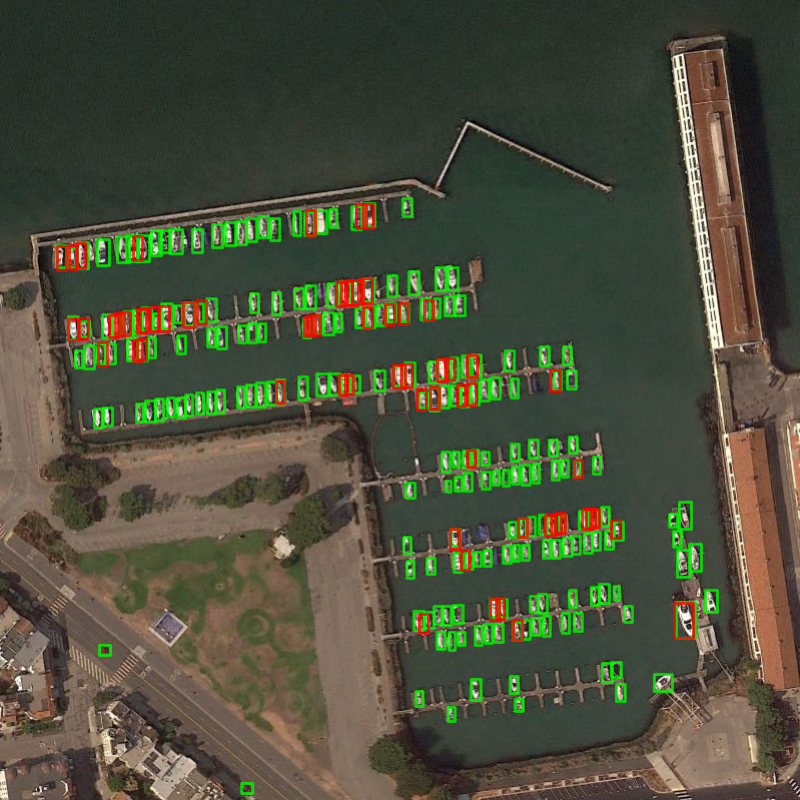}
  \includegraphics[width=0.18\paperwidth]{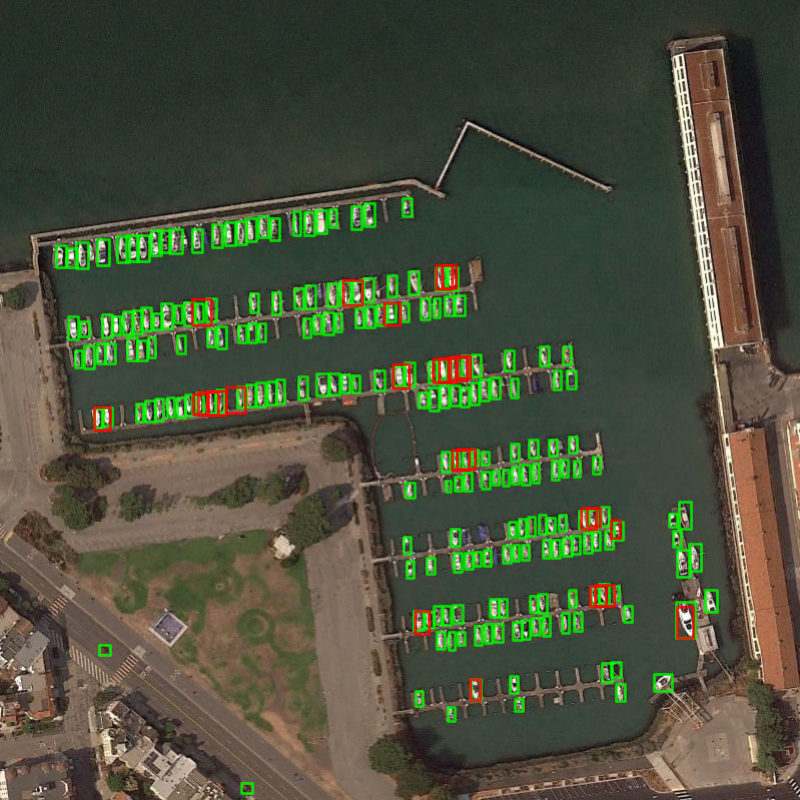}
  \caption{Visualization results on AI-TOD-v2 with RetinaNet. From left to right, they are GCD, NWD, WD, and GIoU. Green boxes represent GT, and red boxes represent predicted boxes. Clearly, GCD shows the best detection performance.}
  \label{Vis_RetinaNet}
\end{figure*}

\subsection{Datasets}

The Aerial Images Tiny Object Detection version 2 (AI-TOD-v2) dataset serves as a bespoke benchmark for the detection of minute objects in aerial imagery. It features 700,621 instances of objects dispersed across eight categories in 28,036 aerial images. The dataset is characterized by a small mean object size of roughly 12.8 pixels, which presents a considerable challenge for detection algorithms. AI-TOD-v2 is a meticulously re-annotated improvement over its predecessor, intending to rectify the prevalent noisy labels and consequently ameliorate the detection capabilities for tiny objects.

VisDrone-2019 comprises both 261,908 video frames and 10,209 still images that represent a diverse compilation of features such as geography (spanning 14 different cities across China), environments (both urban and rural settings), object types (including pedestrians, vehicles, and bicycles), as well as scene density (ranging from sparse to crowded).

The Microsoft Common Objects in Context (MS-COCO) dataset for 2017 is tailored towards tasks including but not limited to object detection, segmentation, key-point identification, and image captioning. This extensive dataset incorporates about 330,000 images, each annotated with 80 categories of objects and 5 descriptive captions, offering an indispensable resource for engineers and researchers engaged in various computer vision endeavors.

\subsection{Ablation Study}
In our ablation studies, RetinaNet and Faster R-CNN were employed as detectors, using the AI-TOD-v2 dataset as the evaluation benchmark. The evaluation metrics adhered to the established protocol of the AI-TOD dataset, encompassing various Average Precision (AP) metrics: $\mathrm{AP}$, $\mathrm{AP_{0.5}}$, $\mathrm{AP_{vt}}$, $\mathrm{AP_{t}}$, $\mathrm{AP_{s}}$, and $\mathrm{AP_{m}}$. Specifically, $\mathrm{AP}$ represents the average precision calculated across multiple IoU thresholds, namely $\{0.5, 0.55, \ldots, 0.95\}$. $\mathrm{AP_{0.5}}$ denotes the AP at an IoU threshold of $0.5$. Additionally, $\mathrm{AP_{vt}}$, $\mathrm{AP_{t}}$, $\mathrm{AP_{s}}$, and $\mathrm{AP_{m}}$ indicate the performance metrics for bounding boxes of very tiny ($2$-$8$ pixels), tiny ($8$-$16$ pixels), and small objects, respectively.

\textbf{Label Assignment and Loss Function.}
We employed Wasserstein Distance (WD), Normalized Wasserstein Distance (NWD), and Gaussian Combined Distance (GCD) as label assignment and regression losses in Faster R-CNN. Table~\ref{tab:Faster} shows that replacing Intersection over Union (IoU) with any of these metrics for label assignment significantly improves performance compared to the baseline. Notably, using GCD alone for label assignment in Faster R-CNN outperforms using WD and NWD for both label assignment and regression loss. Furthermore, employing GCD for both label assignment and regression loss in Faster R-CNN further enhances performance.

\begin{table}[tb]
  \begin{tabular}{c|cc|ccc}  
    Method & Assigning & Bbox Loss & AP & AP$_{50}$ & AP$_{75}$ \\ \hline
    GIoU & & $\checkmark$ & 11.1 & 24.9 & 7.6 \\
    WD & $\checkmark$ & & 18.9 & 46.5 & 11.4 \\ 
    WD & $\checkmark$ & $\checkmark$ & 19.1 & 46.0 & $\mathbf{12.2}$ \\
    NWD & $\checkmark$ & & 17.8 & 44.6 & 9.7 \\
    NWD & $\checkmark$ & $\checkmark$ & 18.4 & 44.1 & 11.4 \\
    GCD & $\checkmark$ & & 19.6 & 48.6 & 11.3 \\
    GCD & $\checkmark$ & $\checkmark$ & $\mathbf{20.1}$ & $\mathbf{48.7}$ & 11.8 \\ \hline                   
  \end{tabular}
  \centering
  \caption{Ablation studies when different metric is applied to multiple modules with Faster R-CNN.}
  \centering
  \label{tab:Faster}
\end{table}

\textbf{Generalization Ability.}
We conducted ablation studies using a wider range of datasets to validate the generalizability of the Gaussian Combined Distance (GCD) 
metric across various dataset scales. Our experiments on the Visdrone-2019 and MS-COCO-2017 datasets demonstrate that Wasserstein Distance 
(WD), Normalized Wasserstein Distance (NWD), and GCD consistently outperform Intersection over Union (IoU) on medium-scale datasets, with GCD 
significantly surpassing both WD\cite{xu2022detecting} and NWD\cite{xue2023visual}. On standard benchmark datasets, 
where WD and NWD performance decline due to scale invariance issues, 
GCD maintains performance comparable to IoU. These findings highlight GCD's robust performance across different scales, offering a distinctive 
advantage not shared by other metrics. Detailed experimental data are presented in Tables~\ref{tab:Vis2019} and~\ref{tab:coco2017}.

\begin{table}[tb]
  \begin{tabular}{l|ccccccc}
    Loss & AP & AP$_{50}$ & AP$_{75}$ & AP$_{vt}$ & AP$_{t}$ & AP$_{s}$ & AP$_{m}$ \\ \hline
    GIoU & 7.6 & 12.6 & 7.9 & 0.0 & 0.4 & 1.6 & 13.9 \\
    WD & 8.0 & 14.2 & 8.2 & 0.1 & 0.4 & 1.7 & 16.1 \\
    KLD & 7.8 & 13.5 & 8.0 & 0.0 & 0.5 & 1.2 & 15.0 \\
    NWD & 7.9 & 14.7 & 7.9 & 0.1 & 0.3 & 1.4 & 15.5 \\
    GCD & $\mathbf{8.7}$ & $\mathbf{15.4}$ & $\mathbf{8.6}$ & $\mathbf{0.2}$ & $\mathbf{0.6}$ & $\mathbf{2.0}$ & $\mathbf{16.8}$ \\ \hline
  \end{tabular}
  \centering
  \caption{Generalization Ability ablation studies on VisDrone-2019 with RetinaNet.}
  \centering
  \label{tab:Vis2019}
\end{table}

\begin{table}[tb]
  \begin{tabular}{l|ccc}
    Loss & AP & AP$_{50}$ & AP$_{75}$ \\ \hline
    GIoU & 36.7 & 57.1 & 39.8 \\
    WD & 31.5$\mathbf{(-5.2)}$ & 50.7$\mathbf{(-6.4)}$ & 33.9$\mathbf{(-5.9)}$ \\
    NWD & 34.6$\mathbf{(-2.1)}$ & 53.2$\mathbf{(-3.9)}$ & 37.2$\mathbf{(-2.6)}$ \\
    GCD & 36.6$\mathbf{(-0.1)}$ & 57.2$\mathbf{(+0.1)}$ & 39.5$\mathbf{(-0.4)}$ \\ \hline
  \end{tabular}
  \centering
  \caption{Generalization Ability ablation studies on MS-COCO-2017 with Faster R-CNN.}
  \centering
  \label{tab:coco2017}
\end{table}

\subsection{Comparison of Peer Methods} 
Table~\ref{tab:SotaRetina} presents the experimental results of various metrics employed as regression losses in RetinaNet on AI-TOD-V2. The data clearly demonstrate that when used as regression losses in RetinaNet, KLD\cite{yang2021learning} slightly outperforms GIoU but remains inferior to DIoU. Notably, the WD-based loss significantly surpasses all IoU-based losses. Most importantly, our proposed GCD method comprehensively enhances the WD-based approach, ultimately achieving state-of-the-art (SOTA) performance.

\begin{table}[tb]
  \begin{tabular}{l|ccccccc}
    Loss & AP & AP$_{50}$ & AP$_{75}$ & AP$_{vt}$ & AP$_{t}$ & AP$_{s}$ & AP$_{m}$ \\ \hline
    GIoU & 6.8 & 17.9 & 4.1 & 2.6 & 8.3 & 7.7 & 23.4\\
    DIoU & 6.9 & 19.5 & 3.6 & 3.8 & 7.3 & 8.4 & 23.4\\
    KLD & 7.3 & 20.0 & 4.1 & 3.2 & 7.4 & 10.8 & 23.7\\
    WD & 9.1 & 24.2 & 4.9 & 2.2 & 8.4 & 14.9 & 25.4\\
    NWD & 8.0 & 21.0 & 4.4 & 2.7 & 8.3 & 13.0 & 25.1\\
    GCD & $\mathbf{11.5}$ & $\mathbf{31.2}$ & $\mathbf{5.7}$ & $\mathbf{3.6}$ & $\mathbf{9.7}$ & $\mathbf{16.0}$ & $\mathbf{28.5}$\\ \hline
  \end{tabular}
  \centering
  \caption{Quantitative comparison of different regression loss on AI-TOD-v2 with RetinaNet.}
  \centering
  \label{tab:SotaRetina}
\end{table}

\section{Discussions}
We discuss a fundamental limitation of the Wasserstein distance: its inherent property of independently optimizing centers. 
This characteristic causes detectors utilizing the Wasserstein distance to lack both scale invariance and precision. 
To address this limitation, we introduce the Gaussian Combined Distance (GCD), a novel metric designed with integrated optimization features to 
enhance the detector's capability to identify small objects while maintaining universality. Empirical evidence demonstrates that our approach 
significantly improves the detector's ability to detect tiny objects, achieving state-of-the-art performance on the AI-TOD-v2 dataset and exhibiting 
robust performance on general datasets—attributes that current metrics do not consistently provide.

\subsection{Expectation}
We have verified the exceptional performance of the Gaussian Combined Distance (GCD) primarily in the realm of horizontal detection. The joint optimization characteristic of the Kullback-Leibler Divergence (KLD)\cite{yang2021learning} has been shown to effectively enhance detection performance in rotational target detection. Consequently, GCD, with its similar properties, may offer unique advantages in rotational target detection with minimal configuration adjustments.

\bibliographystyle{IEEEtran}
\bibliography{references}{}

\end{document}